\documentclass{article}
\usepackage{spconf,amsmath,graphicx}
\usepackage{color}
\usepackage{amsfonts,amsthm}
\usepackage{multirow}
\usepackage{lipsum}

\newcommand\blfootnote[1]{%
	\begingroup
	\renewcommand\thefootnote{}\footnote{#1}%
	\addtocounter{footnote}{-1}%
	\endgroup
}

\title{MCR-Net: A Multi-Step Co-Interactive Relation Network for Unanswerable Questions on Machine Reading Comprehension}

\name{Wei Peng \textsuperscript{1,2}, Yue Hu\textsuperscript{1,2}\sthanks{Corresponding author. E-mail: huyue@iie.ac.cn}, Jing Yu\textsuperscript{1,2}, Luxi Xing\textsuperscript{1,2}, Yuqiang Xie\textsuperscript{1,2}, Zihao Zhu\textsuperscript{1,2}, Yajing Sun\textsuperscript{1,2}}
\address{\textsuperscript{1}Institute of Information Engineering, Chinese Academy of Sciences, China \\
\textsuperscript{2}School of Cyber Security, University of Chinese Academy of Sciences, China}
\begin{document}
%

\maketitle

\begin{abstract}
Question answering systems usually use keyword searches to retrieve potential passages related to a question, and then extract the answer from passages with the machine reading comprehension methods. However, many questions tend to be unanswerable in the real world. In this case, it is significant and challenging how the model determines when no answer is supported by the passage and abstains from answering. Most of the existing systems design a simple classifier to determine answerability implicitly without explicitly modeling mutual interaction and relation between the question and passage, leading to the poor performance for determining the unanswerable questions. To tackle this problem, we propose a Multi-Step Co-Interactive Relation Network (MCR-Net) to explicitly model the mutual interaction and locate key clues from coarse to fine by introducing a co-interactive relation module. The co-interactive relation module contains a stack of interaction and fusion blocks to continuously integrate and fuse history-guided and current-query-guided clues in an explicit way. Experiments on the SQuAD 2.0 and DuReader datasets show that our model achieves a remarkable improvement, outperforming the BERT-style baselines in literature. Visualization analysis also verifies the importance of the mutual interaction between the question and passage.
\end{abstract}
\section{Introduction}
\label{sec:intro}
\blfootnote{
	Copyright 2021 IEEE. Published in ICASSP 2021 - 2021 IEEE International Conference on Acoustics, Speech and Signal Processing (ICASSP), scheduled for 6-11 June 2021 in Toronto, Ontario, Canada. Personal use of this material is permitted. However, permission to reprint/republish this material for advertising or promotional purposes or for creating new collective works for resale or redistribution to servers or lists, or to reuse any copyrighted component of this work in other works, must be obtained from the IEEE. Contact: Manager, Copyrights and Permissions / IEEE Service Center / 445 Hoes Lane / P.O. Box 1331 / Piscataway, NJ 08855-1331, USA. Telephone: + Intl. 908-562-3966.}
Machine Reading Comprehension (MRC) is a long-term goal in natural language processing, which aims to teach machines to understand the given passage and answer questions \cite{DBLP:conf/nips/HermannKGEKSB15}. A machine reading comprehension system should not only answer questions
when possible, but also determine when no answer is supported by the paragraph and abstain from answering. In the extractive MRC task, most of methods \cite{DBLP:conf/iclr/Wang017a,DBLP:conf/iclr/SeoKFH17,DBLP:conf/aaai/YanXWBZZSWWC19} make an assumption that the correct answer is guaranteed to exist in the context passage. Therefore, models only need to select the span that seems most related to the question, ignoring to check whether the answer is actually entailed by the text. So they are still far from true language understanding and are not capable of determining unanswerable questions \cite{DBLP:conf/acl/RajpurkarJL18,DBLP:conf/acl/HeLLLZXLWWSLWW18}, which is much more challenging in practice.

\begin{figure}[tp]
	\framebox{
		\parbox{0.45\textwidth}{
			\small
			\textbf{Article:} Endangered Species Act \newline
			\textbf{Passage:}  \dots Other legislation followed, including the Migratory Bird Conservation Act of 1929 \dots These \textcolor{blue}{later laws} had a low cost to society---the species were relatively rare---and \textcolor{blue}{little opposition} was \textcolor{blue}{raised}. \newline
			\textbf{Question:} \textit{Which laws
				faced \textcolor{blue}{significant opposition}?} \newline
			\textbf{Baseline (plausible answer):} \textit{\textcolor{red}{Later laws}.} \newline
			\textbf{MCR-Net:} \textcolor{green}{None}.}
	}
	\caption{An unanswerable question in SQuAD 2.0,
		along with plausible (but incorrect) answers.
		Relevant clues are shown in \textcolor{blue}{blue}.} 
	\label{fig:exp0}
\end{figure}

In recent years, some initial work \cite{DBLP:conf/acl/GaoDLS18,DBLP:conf/acl/GardnerC18} include that of which uses a classifier with a single fully connected layer to predict the probability of unanswerable questions. However, these studies lacks the explicit process of interaction. \cite{DBLP:conf/aaai/HuWPHYL19} proposes a verifiable system to verify if the candidate answer is actually supported by its surrounding sentence. However, the model uses the plausible answer which may be wrong to make an inference. 
The above methods lack precise positioning and analysis of the key clues that would be helpful to predict unanswerable questions.
Referring to the example in Fig. 1, \textit{later laws} and \textit{little opposition} are key clues in the passage, model can reject to answer the question if it can locate the key clue \textit{little opposition} and then infer that \textit{little opposition} is conflict with \textit{significant opposition} in the question.
\begin{figure*}[t]
	\centering
	\includegraphics[width=0.8\textwidth]{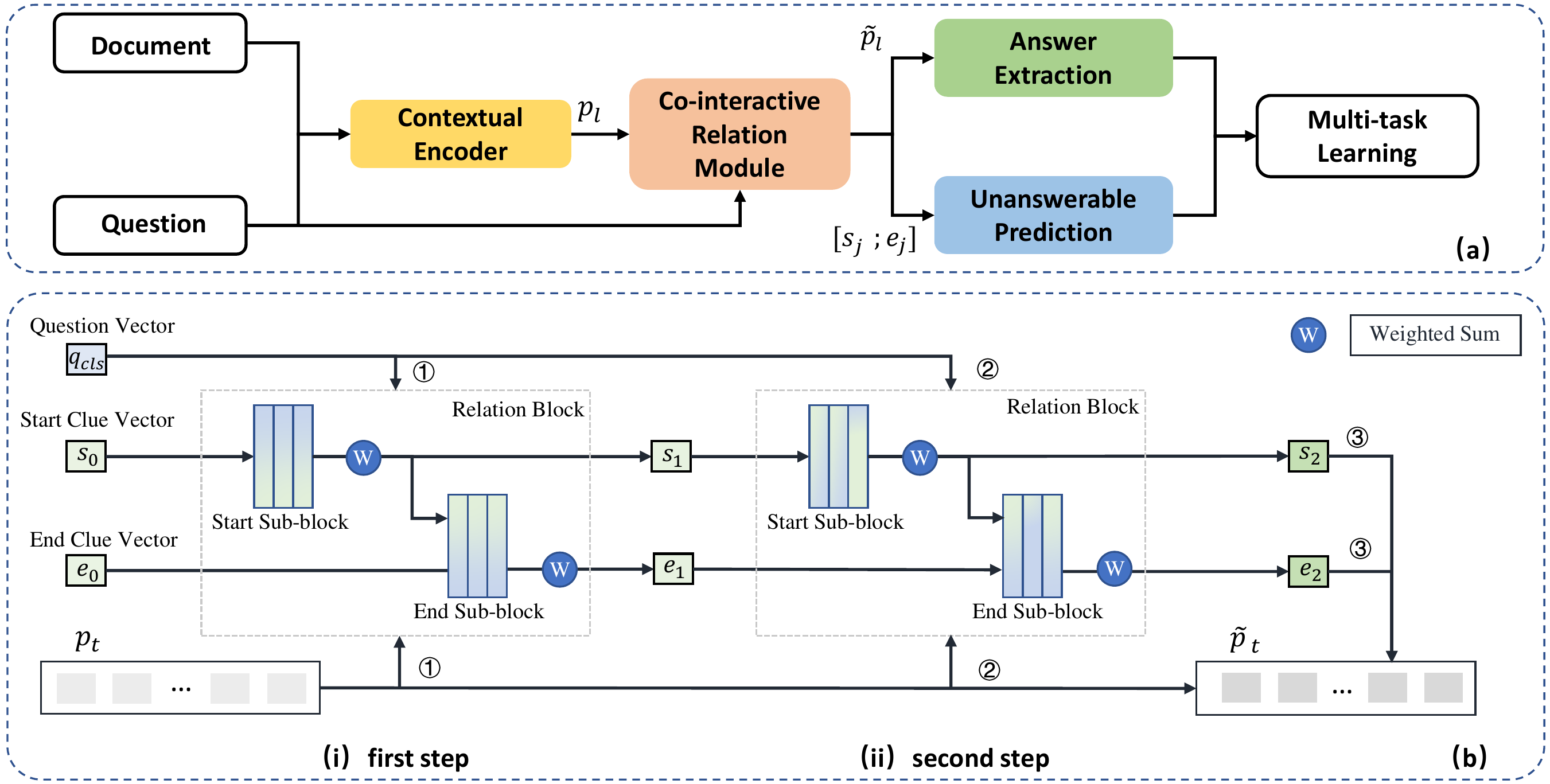} 
	\caption{(a) is the overview of our MCR-Net which consists of the Contextual Encoder, Co-Interactive Relation Module and Answer Predictor. (b) illustrates the details of the Co-Interactive Relation Module. \textcircled{\small{n}} indicates the order of processes.}
	\label{fig0}
\end{figure*}
In this paper, we propose a \textbf{Multi-Step Co-Interactive Relation Network (MCR-Net)} which explicitly models the mutual interaction between the question and passage and locates key clues from coarse to fine for determining unanswerable questions. Specifically, we first consider the pre-trained language model as an encoder to obtain contextual representations of the question and passage. Then, the representations are fed into the \textbf{Co-Interactive Relation Module} to locate key clues for subsequent answer predictor.
The important observation is the significant performance obtained from MCR-Net; it not only achieves the high performance on two datasets but also improves the Recall of unanswerable questions.

\section{METHODOLOGY}
\label{sec:METHODOLOGY}

As shown in Fig. \ref{fig0} (a), the proposed model consists of three modules. First, the Contextual Encoder obtains the contextual representations of passage and question by pre-trained language model \cite{Vaswani2017AttentionIA,DBLP:journals/corr/abs-1909-11942,DBLP:journals/corr/abs-1907-11692} such as RoBERTa and ALBERT. Then, the Co-Interactive Relation Module explicitly models the mutual interaction between the question and passage and locate key clues from coarse to fine. Finally, precise clues are utilized in Answer Predictor to make subsequent predictions.
\subsection{Contextual Encoder}
Considering the strong performance of pre-trained language models, we use them as the Contextual Encoder. Given a question $X^Q = \{x^q_0, \ldots, x^q_{M-1}\}$ and a passage $X^{P} = \{x^{p}_0, \ldots, x^{p}_{N-1}\}$, following the BERT \cite{DBLP:conf/naacl/DevlinCLT19}, the total length of the input is $L = (M + N + 3)$, $M$ and $N$ are the length of the passage and question. To reread the question to locate question-relevant clues, we obtain the representation of the pure question ${q_{cls}}$ for a special token $[CLS]$ in Eqn.2, as:
\begin{align}\label{eq:0}
p_i \!=\! {\rm BERT}({[CLS]}, &x^q_i, {[SEP]}, x^p_i, {[SEP]})  \\
q_j \!=\! {\rm BERT}(&{[CLS]}, x^q_j, {[SEP]})
\end{align}
\subsection{Co-Interactive Relation Module}
The proposed Co-Interactive Relation Module is responsible for explicitly modeling the mutual interaction and relation information between the question and passage and capturing more accurate question-relevant clues. 
Specifically, it continuously integrates and fuses history-guided mutual relations and current-query-guided clues in each step of the interaction.
\\
\textbf{Relation Block.} As shown in Fig. \ref{fig0} (b), the module contains a stack of Relation Blocks, each of which can be stacked to perform multi-step interaction for better utilizing the mutual relation and knowledge. Motivated by pointer network\cite{DBLP:conf/nips/VinyalsFJ15}, the blocks consist of two sub-blocks which interact together to obtain clues fragments. And the calculation of the end sub-block considers the result of the start sub-block. $p_l$ and $q_{cls}$ are regarded as the information flow of the module to gather information from the question and passage multiple times. $ s_j $ and $ e_j $ are the start and end clue vectors in the $j^{th}$ interactive step. The clue vectors fuse all possible clues based on relevance to the question. When $j=$ 0, we use a randomly initialized vector $s_0$ and $e_0$ as start and end clue vectors.

In the \textbf{Start Sub-block}, the module first concatenates $s_j$ and $p_t$ to enhance the representation of the passage $p_t$, as:
\begin{equation}\label{equ:3}
\hat{p}^j_t = {\rm ReLU}({\rm Linear} ([s_j; p_t]))
\end{equation}
where $j = \{0,1,2...\}$ means the $j^{th}$ interactive step, [;] is vector concatenation across row.

To obtain the question-relevant clues, the module rereads the question by using $q_{cls}$. Then it computes a start probability distribution $\alpha^j$, as:
\begin{equation}\label{equ:4}
\alpha^j_t = {\rm Softmax}({q_{cls}}^\top \hat{p}^j_t)
\end{equation}
where $\alpha^j_t \in \mathbb{R}^{L}$, indicating what information the question is more concerned about.

Given the $\hat{p}^j_t$ in Eqn. \ref{equ:3} and $\alpha^j$ in Eqn. \ref{equ:4}, the weighted sum operation will be utilized to obtain the updated start clue vector $s_{j+1}$ which is a guider for the latter interactive.
\begin{equation}\label{equ:5}
s_{j+1} = \sum_{t}\alpha_t^j \hat{p}^j_t
\end{equation}

After calculating the probability distribution $\alpha_t^j$ and the clue vector $s_{j}$ of the start position, we consider the \textbf{End Sub-block}. Because of the temporal relationship, we introduce $s_j$ into the calculation of $e_{j+1}$ to further capture the mutual information. Similarly, the ending probability distribution $\beta^j$ and the end clue vector $e_j$ are yielded as:
\begin{align}
\hat{p}^j_t = {\rm ReLU}({\rm Linear} ([s_{j+1} &; e_j ; p_t]))	\\
\beta^j_t = {\rm Softmax}({\rm Linear}(&{q_{cls}}^\top \hat{p}^j_t))	\label{equ:6} \\
e_{j+1} = \sum_{t} \beta_{t}^j \hat{p}^j_t&	
\end{align}
where $s_{j+1}, e_j \in \mathbb{R}^{h}$ and $\beta^j \in \mathbb{R}^{L}$.

To further interact and locate the key clues from coarse to fine, the model iteratively updates the start and end clues vectors $s_{j-1}$,\, $e_{j-1}$. 
The $s_{j-1}$ and $e_{j-1}$ of the current block are regarded as the initial states of the subsequent block. Circularly, final states $s_j$ and $e_j$ which fuse the mutual interaction and related information are utilized to better capture the semantic relatedness, as $\tilde{p_t} = W^\top_{g} [{p_t; s_{j}; e_{j}}]$. 

\subsection{Answer Predictor}
\textbf{Answer Possibility Classifier.} In order to enable the model to determine whether the question can be answered based on key clues, we utilize $s_{j}$ and $e_{j}$ for binary classification, which is different from previous methods. It is unanswerable when the score exceeds a threshold.
\begin{align}\label{equ:10}
{\rm{score}}_i = W^\top_{s} [s_{j}; e_{j}],  \quad \hat{s}_{i} = \sigma({\rm{score}}_i) 
\end{align}
where $W_{s} \in \mathbb{R}^{2h}$, 
$\hat{s}_{i}$ is a scalar which means the unanswerable score of the $i^{th}$ question.

The average cross-entropy loss among all the questions is optimized as:
\begin{equation} \label{equ:11}
\mathcal{L}_{ans}=-\frac{1}{N}\sum_{i=1}^{N}[\hat{y} \log \hat{s}_{i} +(1-\hat{y}) \log (1-\hat{s}_{i})]
\end{equation}
where $N$ is the total number of the questions, $\hat{y} \in \{0,1\}$ denotes a label, $\hat{y} = 1$ means the $i^{th}$ question is unanswerable.

\noindent
\textbf{Answer Extractor.} Following \cite{DBLP:conf/acl/YangM17}, we consider $\tilde{p_0}$ as a sentinel to indicate the question is unanswerable. The answer span is calculated by predicting the start and the end indices. We compute the final probability $\alpha^j$ and $\beta^j $ of each word to be the start or end position, as:
\begin{equation}\label{equ:8}
\gamma^j_t = {\rm Softmax}[(W^\top_{c} \tilde{p_t})], \quad \eta^j_t = {\rm Softmax}[(W^\top_{e} \tilde{p_t})]
\end{equation}
where $W_{c}$, $W_{e}$ $\in \mathbb{R}^{h}$. $\gamma^j$, $\eta^j$ $\in \mathbb{R}^{L}$. $\gamma_{0}^j$ and $\eta_{0}^j$ denote the score of sentinel that hints non-answerability.

The $\mathcal{L}_{span}$ loss function is to minimize the sum of the negative log probabilities of the true start and end positions. $s$ and $e$ represent the true start and end positions of the answer.
\begin{equation}\label{equ:9}
\mathcal{L}_{span} = -\frac{1}{N}\sum^N_i \log(\gamma^j_s) + \log(\eta^j_e)
\end{equation}

%

\subsection{Joint Learning}
We combine the above two loss functions as the training loss:
\begin{equation} \label{equ:13}
\mathcal{L(\theta)}=\lambda_1\mathcal{L}_{span}+\lambda_2\mathcal{L}_{ans}
\end{equation}
where $\theta$ is the all learnable parameters, and $\lambda_1$ and $\lambda_2$ are two hyper-parameters for controlling the weight of the rest tasks.


\section{Experiments}
\label{sec:pagestyle}
\begin{table}[t]
	\centering
	\resizebox{0.75\columnwidth}{!}{
		\begin{tabular}{lcc}
			\hline
			\textbf{Model}                         & \textbf{Exact Match(\%)} & \textbf{F1(\%)} 	\\
			\hline	
			DocQA \cite{DBLP:conf/acl/GardnerC18}   & 61.90                & 64.80           \\
			Unet\cite{DBLP:journals/corr/abs-1810-06638}						   & 70.30                & 74.00             \\
			RMR + Answer Verifier \cite{DBLP:conf/aaai/HuWPHYL19}						   & 72.30                & 74.80              \\
			Relation Network \cite{DBLP:journals/corr/abs-1910-10843}						   & 79.20                & 82.60             \\
			SG-Net	\cite{DBLP:journals/corr/abs-1908-05147}					   & 85.10                & 87.90              \\
			\hline
			ALBERT-base                            & 76.05               & 80.07             \\
			{MCR-Net-base} (ours)		   		   & {78.25}      & {81.47}    \\
			\hline
			ALBERT-xxlarge                         & 84.78         &  88.08           \\
			\textbf{MCR-Net-xxlarge} (ours) 		   & \textbf{85.63}      & \textbf{88.72}      \\
			\hline
			Human Performance                      & 86.30               & 89.00          \\                
			\hline
	\end{tabular}}
	\caption{\label{tab:0} Exact Match (EM) and F1 scores (\%) on SQuAD 2.0 development set for single model.}
\end{table}

\begin{table}[t]
	
	\centering
	\resizebox{0.75\columnwidth}{!}{
		\begin{tabular}{lcc}
			\hline
			\textbf{Model}                         & \textbf{ROUGE-L(\%)} & \textbf{BLEU4(\%)} \\
			\hline	
			Match-LSTM \cite{DBLP:conf/iclr/Wang017a}    & 39.00                & 31.80                 \\
			BiDAF \cite{DBLP:conf/iclr/SeoKFH17}    & 39.20                &  31.90                \\
			V-Net \cite{DBLP:conf/acl/WuWLHWLLL18}    						   & 44.18                & 40.97            \\
			\hline
			RoBERTa-base                        & 48.35               & 47.00                \\
			{MCR-Net-base} (ours) 			   & {50.16}      & 48.64       \\
			\hline
			RoBERTa-large                        & 49.82               & 48.22                \\
			\textbf{MCR-Net-large} (ours)			   & \textbf{50.82}      & \textbf{49.18}       \\
			\hline
			Human Performance                      & 56.10               & 57.40               \\                
			\hline
	\end{tabular}}
	\caption{\label{tab:1} Performance on the DuReader 2.0 test set.}
\end{table}
\begin{figure*}
	\centering
	\includegraphics[width=0.7\linewidth]{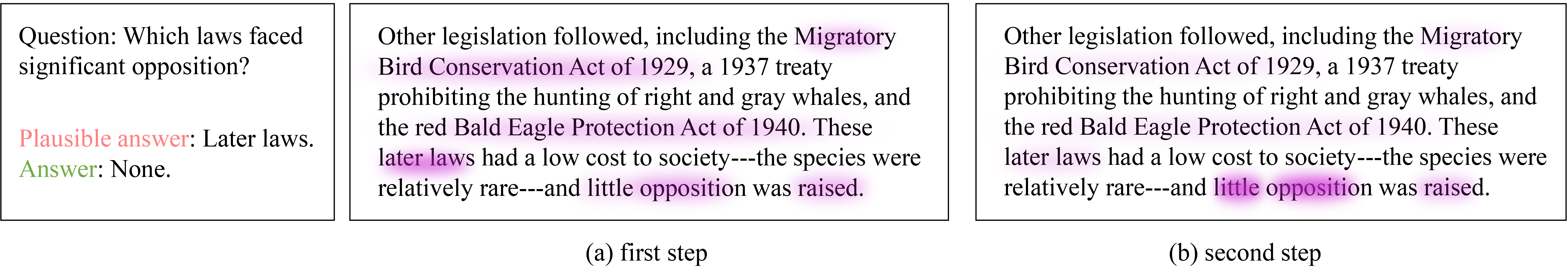}
	\caption{The two subplots show the attention ($\alpha$ and $\beta$) of the first step and second step in the co-interactive relation module respectively. Apparently, the keywords \textit{little opposition} in the passage is highlighted in (b), and the \textit{later laws} is ignored in the second step, which is exactly the wrong answer.}
	\label{fig:visualization2}
\end{figure*}
\subsection{Experimental Setting}
\noindent
\textbf{Datasets \& Evaluation Metrics.} SQuAD2.0 \cite{DBLP:conf/emnlp/RajpurkarZLL16} (contained 150k questions) and DuReader \cite{DBLP:conf/acl/HeLLLZXLWWSLWW18} (contained 300k questions) datasets provide unanswerable questions which contain more than  50,000 (33\%) and 16,890 (5.33\%), respectively. DuReader is a generative task whose answers are manually generated. For fair comparision, we regard it as extractive tasks like most works \cite{DBLP:conf/acl/WuWLHWLLL18} did. Another difference is that SQuAD2.0 is an English dataset while DuReader is a Chinese dataset, aiming to vefiry the generalization of our model. As for the evaluation metric, we use EM \cite{DBLP:conf/emnlp/RajpurkarZLL16}, F1 \cite{DBLP:conf/emnlp/RajpurkarZLL16} and ROUGE-L \cite{lin2004rouge}, BLEU-4 \cite{DBLP:conf/acl/PapineniRWZ02} for the two datasets.

\noindent
\textbf{Implementation details.} The BERT-style baselines have the same hyper parameters as \cite{DBLP:journals/corr/abs-1909-11942,DBLP:journals/corr/abs-1907-11692}. The best pre-trained language model RoBERTa and ALBERT are used for Chinese and English datasets, respectively. We use Adam optimizer \cite{DBLP:journals/corr/KingmaB14} for training, with a start learning rate of 3e-5 and mini-batch size of 32. The number of epochs for SQuAD 2.0 and DuReader is set to 4 and 2, respectively. The hyper-parameters in the loss are $\lambda_1$ = 0.7, $\lambda_2$ = 0.3. And the unanswerable thresholds of the SQuAD 2.0 and DuReader datasets are set to 0.3 and 0.2. 

\begin{table}[!]
	\centering
	\resizebox{0.7\columnwidth}{!}{
		\begin{tabular}{l|lccc}
			\hline
			& \textbf{Model} & \textbf{F1(\%)} & \textbf{R(\%)} &  \textbf{P(\%)}\\
			\hline
			\multirow{2}{*}{\textbf{SQuAD 2.0}}    & ALBERT-base    & 82.37            & 77.00            & \textbf{88.56}               \\ 
			&\multicolumn{1}{l}{MCR-Net-base}         & \textbf{83.97}          & \textbf{81.22}           &     86.83       \\
			\hline
			\multirow{2}{*}{\textbf{DuReader}} & RoBERTa-base   	& 19.35          & 11.87          & \textbf{52.38}          \\
			&\multicolumn{1}{l}{MCR-Net-base}         & \textbf{34.22}         & \textbf{28.04}          & 43.91   \\
			\hline      
	\end{tabular}}
	\caption{\label{tab:4} Precision (P), Recall (R) and F1 of unanswerable questions on the two datasets for baselines and MCR-Net.}
\end{table}

\subsection{Experimental Results}
Due to the limited resources that training the large model would take, we just use the large pre-trained language model in state-of-the-art comparison, while other results are based on the base model. We consider different steps of the interaction, the model performs best in the second step. So the results are reported with $step=$ 2.

\noindent
\textbf{State-of-the-art Comparison.} The main results are shown in Table \ref{tab:0} and Table \ref{tab:1}. The proposed model consistently outperforms the previous methods and baseline models on two datasets. In SQuAD 2.0, \textbf{2.20\%} gain on EM and \textbf{1.40\%} gain on F1 on the ALBERT-base model, as well as improving on the ALBERT-xxlarge model. In DuReader dataset, our model can also make decent improvements. This result illustrates the effectiveness of explicitly modeling the interaction between the passage and question in a multi-step mode, and the performance can be boosted through this mutual information.

\noindent
\textbf{Performance of Unanswerable Questions.} We evaluate the capability of the model to predict answerability in this section, results are described in Table \ref{tab:4}. The recall has improved significantly by \textbf{4.22\%} and \textbf{16.17\%} on two datasets. It can be concluded that our model is more capable of finding unanswerable questions comparing with simple classifiers. And we explain why the question can not be answered by introducing the co-interactive relation module to explicitly locate key clues. Despite the decline in Precision, F1 score still has improved a lot. Another interesting finding is that the Recall on two datasets differs significantly. We make statistics of the number of unanswerable questions on two test results, and find that the number of unanswerable questions on the DuReader is really low (5.33\%). The analysis demonstrates that our method is more robust, however, the simple classifier has a poor performance on class imbalance problems.

\noindent
\textbf{Ablation Study.} To get a better insight into our MCR-Net, we run the ablations 
on SQuAD 2.0 and DuReader, which is shown in Table 4. The stacked relation blocks make a contribution to the overall performance, which confirms our hypothesis that explicitly modeling mutual interaction and relation between the question and passage is important.

\begin{table}[!]
	\centering
	\resizebox{0.8\columnwidth}{!}{
		\begin{tabular}{l|cc|cc}
			\hline
			\multicolumn{1}{c|}{\multirow{2}{*}{\textbf{Model}}} & \multicolumn{2}{c|}{\textbf{SQuAD2.0}}       & \multicolumn{2}{c}{\textbf{DuReader}}       \\ \cline{2-5} 
			\multicolumn{1}{c|}{}	& \textbf{EM}	& $\Delta$	& \textbf{R-L} & $\Delta$                \\ \hline
			\textbf{Complete Model}	& \textbf{78.25}       & \textbf{-}	& \textbf{50.16}	& \textbf{-}           \\	\hline
			- Relation Block	& 76.93	& -1.32	& 49.28	& -0.88                \\
			- Stacked Relation Blocks	& 76.05	& -2.20	& 48.35	& -1.81	\\
			\hline
		\end{tabular}}
	\caption{\label{tab:3} Ablation study on model components. }
\end{table}

\subsection{Interpretability Study}
To demonstrate how the co-interactive relation module works when locating key clues, we conduct an interpretability study with the same example in Sec. 1. The attention maps of the first step and second step in the module are shown in Fig. \ref{fig:visualization2}. In the first step, the module attends some question-relevant clues such as \textit{later laws}, \textit{opposition} and so on. The plausible answer \textit{later laws} is highlighted which is the wrong answer. And the module ignores the key clue \textit{little opposition}. In Fig. \ref{fig:visualization2} (b), we conclude that clues go from coarse to fine, and some coarse clues have been ignored, such as \textit{Migratory Bird Conservation Act of 1929}. The wrong answer \textit{later laws} has been corrected and the key clue \textit{little opposition} is highlighted through multi-step co-interactive relation operation.

\section{CONCLUSION}
\label{sec:foot}

In this paper, we present the
MCR-Net which explicitly models the mutual interaction between the question and passage and locates key clues from coarse to fine rather than in an implicit way. And we try to explain why this question is unanswerable instead of using a binary classifier that does not know the reason. We show that the proposed MCR-Net is effective and robust, which outperforms the previous methods 
with a single model, as well as improving the Recall and F1. 
\section{Acknowledgments}
This work is supported by the National Natural Science Foundation of China (No.62006222).


\bibliographystyle{IEEEbib}
\bibliography{strings,refs}

\end{document}